# Gender Aware Spoken Language Translation Applied to English-Arabic

Mostafa Elaraby , Ahmed Y. Tawfik, Mahmoud Khaled, Hany Hassan , Aly Osama

Microsoft AI & Research

a-moelar,atawfik,t-makh,hanyh,a-alosam@microsoft.com

**Abstract.** Spoken Language Translation (SLT) is becoming more widely used and becoming a communication tool that helps in crossing language barriers. One of the challenges of SLT is the translation from a language without gender agreement to a language with gender agreement such as English to Arabic. In this paper, we introduce an approach to tackle such limitation by enabling a Neural Machine Translation system to produce gender-aware translation. We show that NMT system can model the speaker/listener gender information to produce gender-aware translation. We propose a method to generate data used in adapting a NMT system to produce gender-aware. The proposed approach can achieve significant improvement of the translation quality by 2 BLEU points.

**Keywords**: Speaker Gender, Gender Information, Gender Aware Translation, Gender Agreement, produce gender, Neural Machine Translation System,

## 1 Introduction

Nearly half the world languages have a grammatical gender system. For native speakers of these languages, violations of gender agreement are associated with a difficulty in comprehension. In one study [1] , gender agreement violations resulted in a delay of 500 to 700 ms. in response time while reading Spanish sentences. A similar study [2], has reached analogous conclusions for spoken language comprehension. These findings suggest that gender agreement violations place an additional cognitive overload on the listener.

In conversational settings, pronouns are frequently used referring to the speaker or addressing the listener(s). Pronominal gender agreement is particularly challenging for machine translation (MT), particularly when the source language does not have gender agreement while the target language does, which is the case for English to Arabic translation. The focus of this paper is to enable a SLT system to produce gender-aware translation for both parties participating in a conversation.

For instance, let us consider a SLT session involving English and French participants. If an English says: "I am certain". The appropriate translation of the adjective "certain" to French depends on the speaker gender since French has a grammatical gender system. For a male speaker the correct translation is "Je suis certain", while "Je suis certaine" is the correct form for a female speaker. Similarly, in Arabic, "I am certain" should be translated to "أنا متأكد" (*?na mt?kd*) or "أنا متأكدة" (*?na mt?kdt*) for a male or female speaker respectively.

The listener's gender would affect the translation as well. Let's consider the translation of "You said it" into Arabic. For a male listener, it should be "أنت قلته" (*?nt qlth*) and for a female listener, the correct translation becomes "أنت قلتيه" (*?nt qltyh*). As the listener is also a speaker in conversational setting, the term "speakers gender agreement" here refers to both "speaker-dependent" and "listener-dependent" gender agreement unless making the distinction is necessary for the clarity of the presentation.

To assess the prevalence of speakers' gender agreement in SLT, we have randomly selected 1000 sentences from the English-Arabic Open-Subtitles data [3]. These sentences were manually analyzed for speaker-dependent or listener-dependent gender agreement. More than half the sample contained at least one form of gender dependency. However, smaller number of sentences, had both speaker and listener dependency. Detailed findings are in Table 1. We also observed that the listener dependency is much more dominating than speaker dependency.

**Table 1.** Gender dependence in 1000 Open-Subtitle sentences.

| Gender Dependence | Percentage of Sentences |
|---|---|
| None | 48.5% |
| Speaker Only | 3.1% |
| Listener Only | 46.9% |
| Both Speaker and Listener | 1.5% |

Fortunately, speaker gender determination from speech has reached high accuracy even for relatively short speech segments [4]. Therefore, we can rely on having this information at runtime. However, training a SLT system would require gender tagged parallel sentences to be able to generate gender-aware translations. This is particularly important in the current pipelined approach to SLT, which combines a speech recognition component followed by machine translation, commonly used in large scale SLT systems. A promising direction is training end-to-end speech to speech translation systems [5] which is trained on source language audio and produces target language audio(or text). In such setting, the speaker's gender information can be easily extracted from the source language audio. However, the listener's gender information would still be required to be able to produce gender-aware SLT.

One of the main challenges in training gender-aware SLT is to find a large gender tagged parallel corpus that has both the speaker's and listener's gender information. To address this challenge, we propose an approach to automatically label a parallel conversational corpus with gender information. Applying this approach to the Open Subtitle data set has produced the training data needed for this work. The proposed approach uses a part-of-speech tagger and a set of rules to automatically tag sentences with speaker and listener genders. The tagged sentences are used to adapt a baseline neural MT system trained using sequence to sequence training with attention. This baseline system is trained using both gender dependent and gender independent sentences, then adapted using the sentences with identified gender dependence.

The main contribution of this paper is twofold: enabling NMT systems to produce gender-aware translation and provide a method to generate the data to achieve that. The remainder of this paper is structured as follows. Section 2 reviews some of the work on speaker gender determination from speech. Section 3 describes the sentence labelling process for speaker gender dependent and listener gender dependent utterance extraction. Section 4 outlines the NMT training and testing used. Section 5 summarizes the experiments we have conducted, and Section 6 concludes the paper.

## 2  Speaker Gender Identification

Humans can easily identify the gender of the speaker from a noticeably short audio segment due to the natural differences in female and male speech generation process. Automatic gender identification helps in improving the accuracy and the robustness of many speech applications such as automatic speech recognition, emotion recognition, content-based multimedia indexing systems, speaker diarization, speaker indexing, human-machine interaction, and voice synthesis.

Gender classification from speech is considered a solved problem on clean and monolingual corpora such as the TIMIT [6] speech corpus or distorted and multilingual corpora such as DARPA RATS [7]. However, differentiating the gender of children from speech is still challenging. In [8], an accuracy of only 85% is reported in a 3-way classification between male, female and child speech.

There are several approaches for identifying gender from speech. One of basic approach is using gender-dependent features such as pitch or fundamental frequency (f0). This approach can reach 100% accuracy on a clean speech dataset like TIMIT with average pitch frequency as a separation criterion [6] which shows that pitch is an important feature for speaker gender classification. The main problem with using pitch is its noticeable deterioration with signal distortion and noise as in telephony speech applications. In addition to that, it shows a lower accuracy with datasets which have a variety of speakers and different languages. Also, the pitch tends to be similar in some female and children as well. Therefore, it is not generally sufficient to use pitch only as a discriminant feature.

Mel-frequency cepstrum coefficients (MFCCs) are commonly used as features in speech, speaker recognition, and audio similarity measures. Most often, MFCCs are the main feature in speech as they capture speech signal changes in time and frequency domain with low computational overhead. Although it is not robust with noisy data, using apocopate frames and normalization of MFCC values modeled by GMM, UBM and SVM show significant gains [9] .

Some Researchers [6] have proposed an effective gender identification system based on identity vector (I-vector) as it retains the main speaker characteristics including speaker's identity, gender, dialect, language, and age. I-vector represents the speech frame with a fixed-size vector, but its dimensionality could be reduced with approaches such as Principal component analysis (PCA) and Linear Discriminate Analysis (LDA). A combined approach of I-vector and LDA achieves 99.9% accuracy on TIMIT and NUST603-2014 datasets [6].

Existing statistical learning methods such as GMM, K-nearest neighbor (KNN), neural networks and support vector machines (SVM) have been explored in gender classification. SVM classifier accuracy reached to 100% on clean speech database and 97.8% on a noisy database [10] . Combination of several types of features such as MFCCs coefficients, pitch, time-domain and frequency domain descriptors as well as using majority voting among (KNN, MLP, SVM, Naive Bayes and Random Forest) shows100% accuracy  [9] on Eustace, a noisy English dataset which indicates a significant result over using SVM only with 95% accuracy rate.

In summary, most recent techniques could identify gender from speech easily, so we can integrate this component in a Spoken Language Translation system to provide the speakers' gender in a two-way conversational setting.

# 3   Extracting Gender Dependent Parallel Sentences

As parallel Arabic English datasets are predominantly in text format. It is still challenging to obtain relevant training data for this work despite the reliability of gender determination from speech. The Egyptian CALLHOME **[5]** parallel dataset has gender labels available for each utterance. However, this dataset uses the Egyptian dialect and is relatively small, on its own, for training a high quality NMT system. Using the parallel English Arabic TED Talks corpus proved to be problematic as well; as the identity of the speakers is provided, thus speakers' gender can be determined. However, there was no meaningful way to represent listeners. Hence, we opted to use the Open-Subtitles data **[3]**.

## 3.1   Filtering Open Subtitles

Despite efforts to accurately align Open-Subtitles data, the subtitles data alignment remains noisy. While some techniques have been proposed to improve the data alignment [11], we opt to limit the data used in this work to a subset with good alignment. The data filtration follows the same approach as [12]. In this approach, the alignment score is used along with source and target word counts, unaligned word ratio, and percentage of one-to-one alignments as features to train a decision forest classifier to differentiate between well aligned sentences and poorly aligned ones based on a set of manually labelled parallel sentence pairs.

## 3.2   Gender Labelling of Training Data

For automatic annotation of the Open-Subtitle dataset with genders of both the speakers and the listeners, an Arabic part-of-speech (POS) tagger[1] provides some meaningful clues to determine these genders. Upon careful examination of the output of the POS tagger on a sample of sentences, we deduced a set of rules to determine the gender information. Table 2 presents these rules. The notation "DoublyTransitiveVB" denotes a verb from a set of transitive Arabic verbs that take two objects and indicate "knowledge" or "transformation" of a property (the second object) about its first object. For example, "علمتك مجتهدا" ("I have known you diligent") would indicate the listener gender as the adjective (JJ) must agree with it. The notation $Pron_O$ and $Pron_S$ denote object and subject pronouns respectively. For example, for the sentence: أنا متأكد, *?na mt?kd* (I am certain-male), the POS tagger labels the adjective as a masculine adjective, and we can infer that the speaker is male as it followed the $Pron_S$ أنا, *?na* (I). Similarly, for a listener, the gender label assigned to an adjective following the pronoun أنت, *?ant* (you) determines the listener 's gender.

VBI here refer to a set of "incomplete verbs" used to describe state like "كان, *kan*, (was); أصبح, *?SbaH* (became), ...etc. Most of the rules determining the speaker's gender rely on adjectives because gender doesn't affect unambiguously other parts of speech. However, for the listener gender, the rules rely on adjectives, verbs and calling particles. Most verbs in second person can indicate the gender of the listener. For example, when a speaker addresses a female saying: "أنت تلعبي", *?nt tlEby* (You are playing), the verb ending differs from a male is addressed: "أنت تلعب", *?nt tlEb* (You are playing).

The calling structures starting with a calling particle like " يا, *ya* (O)", "أيها, *?yha* ", or "أيتها, *?ytha*" can also determine the listener's gender. For example, "يا رجل", *ya ragul* (O man) will allow us to determine that the listener is a male as the word for man is tagged as a male noun. The same applies is the calling preposition was followed by an adjective.

**Table 2.** Gender Labelling Rules.

| Sentence Containing | Listener or Speaker | Gender |
|---|---|---|
| Doubly Transitive VB + Prono. 2$^{nd}$ Pers. Sing JJ | Listener | JJ Gender |
| Pron$_S$. 2$^{nd}$ Pers.Sing JJ | Listener | JJ Gender |
| Pron$_S$. 2$^{nd}$ Pers. Sing. VBI JJ | Listener | JJ Gender |
| VB. 2$^{nd}$ Pers. Sing. Fem. | Listener | Fem. |
| VB. Imperative. 2$^{nd}$Pers.Sing | Listener | VB Gender |
| Call Particle NN or JJ | Listener | NN/ JJ Gender |
| DoublyTransitiveVB+Pron$_O$. 1$^{st}$Pers.Sing JJ | Speaker | JJ Gender |
| Pron$_S$. 1$^{st}$ Pers.Sing JJ | Speaker | JJ Gender |
| Pron$_S$. 1$^{st}$ Pers.Sing VBI JJ | Speaker | JJ Gender |
| VB (No Pron$_O$.2$^{nd}$Pers attached & aligns with a phrase containing "you") JJ | Listener | JJ Gender |
| VB (aligns with a phrase containing "I") JJ | Speaker | JJ Gender |

To evaluate the set of rules, we treated each pair of gender and identity as a class, so we had four classes: Speaker is Male, Speaker is Female, Listener is Male, and Listener is Female. We applied the rules to a random sample of 1000 sentences and calculated precision and recall and obtained the results shown in Table 3.

**Table 3.** P/R for Arabic Only Gender Annotation Rules

| Metric | Male Speaker | Female Speaker | Male Listener | Female Listener |
|---|---|---|---|---|
| Precision | 80.00% | 100% | 63.15% | 93.33% |
| Recall | 19.04% | 25% | 11.65% | 51.85% |

The low recall in Table 3 is due to the ambiguity of some verbal forms. For example, أصبحت, SbaHt could mean either I became, you became, or she became. When used in the first meaning, it could determine the speaker's gender when followed by an adjective, the second would tell us the listener's gender and the third tells nothing about the speaker's or listener's gender. To resolve this ambiguity, we search the aligned English phrase for the pronouns "I", and "you" and thus resolve the ambiguity. After, taking into consideration the aligned pronouns for the ambiguous verbal forms, we get a considerable improvement in both precision and recall as shown in Table 4.

**Table 4.** P/R for Arabic English Annotation Rules

| Metric | Male Speaker | Female Speaker | Male Listener | Female Listener |
| --- | --- | --- | --- | --- |
| Precision | 91.66% | 100% | 92.3% | 95.23% |
| Recall | 52.38% | 50% | 70.27% | 71.42% |

From Table 4, the recall for the speaker classes is still low. As only a small number of sentences are affected by the gender of the speaker in a linguistic way, such as the gender of the adjectives changing, and in the rest of the sentences, the gender of the speaker could only be inferred from a deeper semantic analysis of the sentences.

Consider the sentence: "I consider myself lucky", which has been translated to أعتبر نفسي محظوظ, "*Etbr nfsy mHZuZ*". Here, the gender of the adjective "lucky, محظوظ, *mHZuZ* indicates that the speaker is a male. For a female speaker, the adjective would have taken the female form محظوظة, *mHZuZt*. However, the same "I consider myself" structure becomes more ambiguous if it is followed by a noun as in the following two examples:
- I consider myself an optimist
- I consider myself a victim.

In the first example, the word "optimist" maps to an adjective "متفائل, *mtafa?l*", which would indicate the gender of the speaker. However, in the second example, the word "victim" in Arabic is "ضحية, *DHyt*" which a female noun irrespective of the gender of the speaker.

## 4   Neural Translation and Gender Adaptation

The neural machine translation system is implemented as an attentional encoder-decoder network. The encoder is a bidirectional neural network with LSTM that reads an input sequence $x = (x_1, \ldots, x_m)$ and calculates a forward sequence of hidden states $(\overrightarrow{h_1}, \ldots, \overrightarrow{h_m})$, and a backward sequence $(\overleftarrow{h_1}, \ldots, \overleftarrow{h_m})$. The hidden states $\overleftarrow{h_j}$ and $\overrightarrow{h_j}$ are concatenated to obtain the attention vector $h_j$.

The decoder is a recurrent neural network that predicts a target sequence $y = (y_1, \ldots, y_n)$. Each word $y_i$ is predicted based on a recurrent hidden state $s_i$, the previously predicted word $y_{i-1}$, and a context vector $c_i$. $c_i$ is computed as a weighted sum of the annotations $h_j$.

The weight given to each annotation $h_j$ is computed through an alignment model giving the probability that word $y_i$ is aligned to word $x_i$. The alignment model is a single-layer feedforward neural network that is learned jointly with the rest of the network through backpropagation.

During training all parameters are optimized jointly using Adadelta to maximize the conditional probability of sentence pairs in the training data. At decoding time, one word is predicted using a beam search to score the best translation path.

Using gender annotated data generated in Section 3, along with unannotated data we trained a NMT model with the above architecture having labeled data and large set of unlabeled data.

This trained model was then adapted using the data with gender labels to generate a model that restores gender agreement in the target language. A similar approach was previously used to modulate the politeness of expressions in the translation from English to German [13]. Unlike [13], we trained a base model with all the data that passed the alignment filter; this model is then adapted with the sentences having gender labels. The adaption steps bias the distribution of the gender affected words and promotes its generation for the matching gender of speakers. We use four labels to denote speaker gender information and listener gender information, such that each sentence has at most one speaker gender label and one listener gender label. However, in many cases we had only either the speaker or the listener labeled.

The above translation system was biased toward the gender that is dominant in the input training set, this gender information is lost in the source English data and needs to be restored in the target Arabic language output.

In this work, we use an implementation like [14]. A sequence to sequence encoder-decoder Neural Machine Translation model (NMT) along with attention mechanism which generates a representation capturing the importance of encoded input sentence at the current decoding step to handle large sentences [15].

We would like our translation system to abide to the gender information of both speaker and listener marked as a meta-data input on the source sentence as generated above. The basic idea is to provide the neural network with additional input features or constraints as in [13]. At training time, the correct gender information is provided through the generated data. At test time, we assume that the gender information can be provided by a gender identification system. We add the gender information as a special token at the end of the source text such that the model would be able to learn from that feature and produce the appropriate gender information.

## 5 Experimental Setup

The goal of the experiment described here is to assess the impact of the gender information on the quality of the translation.

### 5.1 Labelled Data

The gender inference rules described in Section 3.2 generated 900k gender tagged sentences when applied to a corpus of 4 million sentences. This corpus was obtained after applying the alignment filter in Section 3.1 to a 6 million sentence Open-Subtitle corpus.

To train our Neural Machine Translation system, we used the whole 4 million sentences minus 1.3K sentences set aside for testing. The gender labels were appended to the English side, and each of these labels had its own separate entry in the English source vocabulary, so that the model can train a word embedding for each one separately.

We trained the model for 7 epochs with a learning rate of 0.001. After the baseline model finished training, we adapted the model using only the 900K sentences that were labelled with a gender. We tested two configurations for the adaptation: the first performed two adaptation epochs at the same learning rate and the second used 10 adaptation epochs at a learning rate of 0.0001 (one tenth of the original learning rate). The first configuration resulted in an overall regression of 0.38 point compared to the base model before adaptation, while improving the BLEU score for gender sensitive sentences by two points. The second configuration performed better in terms of training loss and translation quality (overall BLEU score and score on gender sensitive sentences). The lower learning rate allowed the model to make the necessary adjustment for accounting for gender without disrupting the base model. Therefore, the results reported in the remainder of this section are all obtained using the lower adaptation learning rate.

### 5.2 Triggering Module

A potential mismatch between the training condition and the actual runtime condition can occur as the speaker gender extractor from speech produces the speakers gender labels regardless of the utterance gender sensitivity, while the adaptation data had only gender sensitive sentences. Therefore, an additional classifier prevents the gender tags from affecting gender insensitive utterances.

Given an English sentence, the triggering module determines if the translation should use the base model (for sentences that are not affected by gender) or the adapted one (for gender affected sentences). Following an approach analogous to Section 3.2 above, an English Part-Of-Speech tagger [16] serves as a basis for the NMT model selection rules here.

**Table 5.** Triggering rules for the adapted model.

| Sequences |
|---|
| "I am" RB* JJ |
| "you are" [JJ\|VBG] |
| ^VB |
| "you" VBP |
| "you" JJ |

In Table 5, "RB, JJ, VBG, VBP, VB, MD, and PRP" denote "adverb, adjective, present participle, present tense verb, verb, modal verb, and personal pronoun" respectively. The "*" denote zero or more repetitions, "+" denotes one or more repetitions, and "^" marks the start of a sentence. Using a set of 1000 randomly selected and manually labelled sentences from Open-Subtitles, we evaluated the precision and recall of these rules. As shown in Table 6, these values are relatively high.

**Table 6.** P/R for triggering sequences

| Metric | Score |
|---|---|
| Precision | 95% |
| Recall | 80% |

There are some similarities between these triggering sequences and the rules used for the Arabic sentence labeling. In fact, some of them are POS level translations. For example, the sequence "I am" or "you are" followed by an adjective, maps directly to a labelling rule in Arabic. However, only the Arabic POS tagger can provide the gender labels. Some rules are not as easy to carry over from Arabic, like imperative verbs since they have no clear form, and there is no way to distinguish between them and other forms verbs, unlike Arabic where imperative verbs have a clear form that can be easily determined by the POS Tagger. Often, a verb at the beginning of a sentence is an imperative verb, with the obvious exception of auxiliary verbs in interrogative forms.

The sequences involving "you" followed by a verb, in any form, tend to depend on the listener's gender due to the 2nd person suffixes or prefixes that are gender dependent. A similar pattern is noted for modal verbs, like "can", "may", "must", or "should".

### 5.3 Evaluation

To test the end-to-end system, all 1,300 English sentences in the test set are tagged using the POS tagger. The triggering sequences extracted 300 sentences that would be affected by gender. Then to determine the effect of gender restoration, we compared using the base model (before adaptation) on the whole test set against the proposed approach, where the base model is used for the 1000 sentences that do not match any triggering sequence, and the adapted model for the 300 sentences matching a triggering sequence. The results are as follows using a test set including 300 gender labelled sentences and 1k sentences that are not affected by gender and doesn't have gender labels:

**Table 7.** Bleu Scores

| Model | Full Test Set | Gender Labeled Test Set |
|---|---|---|
| Baseline | 20.49 | 20.04 |
| Proposed | 21.07 | 22.18 |

The proposed approach outperformed the base model for all the search strategies that we tested. An improvement of two BLEU points was observed for the affected sentences.

## 6  Conclusions

This work addresses gender agreement violations as a source of translation errors that has not been addressed previously. Spoken language translation can benefit from the availability of speech and the high accuracy of gender determination from speech in solving this problem.

While the present treatment focused on the English to Arabic pair, the approach can be applied to other languages if we can design a technique to accurately label a sufficiently large set of parallel sentences with gender label.

Gender is only one of the meta information that can be extracted from speech and that may affect spoken language translation. Other demographic aspects such as age, and regional dialect can be effective determined form few seconds of speech. Moreover, other more transient aspects such as emotions, intonation, and pitch could carry meaning that can improve spoken language translation.